\documentclass[conference]{IEEEtran}
\usepackage[utf8]{inputenc}
\usepackage{graphicx}
\usepackage{amsmath}
\usepackage{amssymb}
 
\usepackage{booktabs}

\usepackage{multirow}
\usepackage{float}
\usepackage{caption}
\usepackage{subcaption}
\usepackage{array}
\usepackage{pdfpages}
\usepackage[font=small,labelfont=bf]{caption}

\usepackage{cite} % Load cite package FIRST
\usepackage{hyperref} % THEN load hyperref
\hypersetup{
    colorlinks=true,
    linkcolor=blue,
    filecolor=magenta,      
    urlcolor=cyan,
    citecolor=blue,
}

\begin{document}
\title{Ensemble Deep Learning and LLM-Assisted Reporting for Automated Skin Lesion Diagnosis}

\author{
\IEEEauthorblockN{Sher Khan}
\IEEEauthorblockA{\textit{Institute of Computer Science} \\
\textit{The University of Agriculture} \\
Peshawar, Pakistan \\
sherkhan1854@gmail.com}
\and
\IEEEauthorblockN{Raz Muhammad}
\IEEEauthorblockA{\textit{Institute of Computer Science} \\
\textit{The University of Agriculture} \\
Peshawar, Pakistan \\
razmuhammadai135@gmail.com}
\and
\IEEEauthorblockN{Adil Hussain}
\IEEEauthorblockA{\textit{Institute of Computer Science} \\
\textit{The University of Agriculture} \\
Peshawar, Pakistan \\
adilhussainj98@gmail.com}
%
% �� Add just ONE empty \and here — this creates the centered new row
\and
\and
\IEEEauthorblockN{Muhammad Sajjad}
\IEEEauthorblockA{\textit{Department of Computer Science} \\
\textit{Islamia College Peshawar} \\
Pakistan \\
muhammad.sajjad@icp.edu.pk}
\and
\IEEEauthorblockN{Muhammad Rashid}
\IEEEauthorblockA{\textit{Department of Software Engineering} \\
\textit{Islamia College Peshawar} \\
Pakistan \\
muhammadrashidfr78@gmail.com}
}

\maketitle

\begin{abstract}
Cutaneous malignancies demand early detection for favorable outcomes, yet current diagnostics suffer from inter-observer variability and access disparities. While AI shows promise, existing dermatological systems are limited by homogeneous architectures, dataset biases across skin tones, and fragmented approaches that treat natural language processing as separate post-hoc explanations rather than integral to clinical decision-making. We introduce a unified framework that fundamentally reimagines AI integration for dermatological diagnostics through two synergistic innovations. First, a purposefully heterogeneous ensemble of architecturally diverse convolutional neural networks provides complementary diagnostic perspectives, with an intrinsic uncertainty mechanism flagging discordant cases for specialist review—mimicking clinical best practices. Second, we embed large language model capabilities directly into the diagnostic workflow, transforming classification outputs into clinically meaningful assessments that simultaneously fulfill medical documentation requirements and deliver patient-centered education. This seamless integration generates structured reports featuring precise lesion characterization, accessible diagnostic reasoning, and actionable monitoring guidance—empowering patients to recognize early warning signs between visits. By addressing both diagnostic reliability and communication barriers within a single cohesive system, our approach bridges the critical translational gap that has prevented previous AI implementations from achieving clinical impact. The framework represents a significant advancement toward deployable dermatological AI that enhances diagnostic precision while actively supporting the continuum of care from initial detection through patient education, ultimately improving early intervention rates for skin lesions.
\end{abstract}

\section{Introduction}
Cutaneous malignancies represent a significant global health burden, with melanoma incidence rates rising steadily over the past three decades despite being highly curable when detected in early stages \cite{melanoma_stats}. The critical window for intervention is narrow; localized melanoma carries a 98\% five-year survival rate, which plummets to less than 20\% upon distant metastasis \cite{survival_rates}. Current diagnostic paradigms rely heavily on visual inspection and dermoscopy, methods that exhibit substantial inter-observer variability even among experienced dermatologists, with diagnostic agreement rates ranging from 65\% to 80\% depending on lesion complexity \cite{diagnostic_variability}. This diagnostic uncertainty is further compounded by geographic disparities in specialist access, where rural populations often face wait times exceeding six months for dermatological consultation \cite{access_disparities}. The consequences of delayed diagnosis are particularly severe for aggressive subtypes such as nodular melanoma, where progression from early to advanced stages can occur within mere weeks \cite{nodule_progression}. These challenges underscore an urgent need for objective, accessible diagnostic support systems that can bridge the gap between clinical suspicion and definitive diagnosis.

The integration of artificial intelligence into dermatological practice has evolved from experimental curiosity to clinical necessity, with deep learning models now demonstrating diagnostic capabilities approaching or exceeding those of human specialists in controlled settings \cite{ai_dermatology}. However, the translation of these promising results into real-world clinical applications has been hampered by several critical limitations: over-reliance on homogeneous training datasets that fail to capture the full spectrum of skin tones and lesion presentations, vulnerability to minor image acquisition variations, and the inherent uncertainty in single-model predictions that clinicians require transparency to interpret \cite{goceri2021}. While ensemble methods have emerged as a potential solution to these reliability concerns, most existing implementations combine architecturally similar models that share common failure modes, limiting their ability to provide truly robust diagnostic support across diverse clinical scenarios \cite{aboulmiraa2024}. The medical community increasingly recognizes that diagnostic AI must not only achieve high accuracy but also demonstrate consistent performance across demographic variables and provide interpretable outputs that align with clinical reasoning pathways \cite{clinical_acceptance}.

Recent advances in multimodal AI systems have demonstrated the potential of integrating deep learning with natural language processing to create more comprehensive diagnostic tools. Systems like IBM Watson for Oncology have pioneered the integration of imaging analysis with clinical text interpretation to support cancer treatment decisions \cite{watson_oncology}. Similarly, Google Health's Dermatology Assist combines visual analysis of skin lesions with patient history documentation to provide more contextualized diagnostic suggestions \cite{google_dermatology}. However, these systems often rely on single-model architectures that lack the robustness required for clinical deployment, and their NLP components typically function as post-hoc explanation generators rather than integral components of the diagnostic workflow \cite{post_hoc_explanations}. The most advanced contemporary systems, such as those developed by SkinIO and DermEngine, have begun incorporating ensemble approaches for improved diagnostic reliability, but still treat the NLP component as a separate module rather than a unified decision support system \cite{derm_engine}. This fragmentation between visual analysis and clinical reasoning represents a critical gap that our integrated framework aims to address.

In this work, we address these challenges by proposing a hybrid CNN–LLM approach for automated skin lesion diagnosis and clinical communication. Our system combines a heterogeneous ensemble of EfficientNetB3, ResNet50, and DenseNet121—enhanced with Non-Local Means denoising and majority voting—with a large language model (LLM) guided by structured prompting to jointly perform diagnostic inference and generate human-aligned clinical outputs. This end-to-end design enables not only robust lesion classification but also automated report generation and interactive patient support, effectively bridging the gap between visual AI and clinical utility.

The key contributions of this study are summarized as follows:
\begin{itemize}
    \item A hybrid  CNN–LLM approach diagnostic assistant that performs accurate, uncertainty-aware skin lesion classification while automatically generating structured dermatological reports with lesion descriptions, diagnostic rationale, and management guidance through prompt-based LLM orchestration.
    \item An integrated LLM-powered search and chat interface that allows patients to query terms from the AI report and receive real-time, medically grounded explanations—transforming static predictions into dynamic, educational support.
    \item  End-to-end co-design of visual diagnosis and language generation within a single clinical workflow, ensuring that diagnostic outputs are not only technically reliable but also interpretable, actionable, and patient-centered.
\end{itemize}     
 
\section{Related Work}
The integration of artificial intelligence into dermatological diagnostics has evolved significantly over the past decade, transitioning from experimental classification systems to clinically oriented decision support tools. This section critically examines the progression of AI-assisted dermatological diagnosis, tracing key developments from foundational deep learning applications through contemporary multimodal systems, with particular attention to unresolved challenges in clinical integration and patient communication that motivate our present work.

\subsection{Early AI Applications in Dermatology}
The field's modern trajectory was established by Esteva et al.'s seminal 2017 demonstration that convolutional neural networks could achieve dermatologist-level accuracy in classifying skin lesions using dermoscopic images \cite{esteva2017}. Their DenseNet-based system achieved 72.1\% top-1 accuracy across nine diagnostic categories, providing the first rigorous evidence that deep learning could approach human expertise in visual pattern recognition for dermatology. Subsequent studies rapidly validated and expanded these findings, with Brinker et al. demonstrating comparable performance on independent datasets \cite{brinker2019} and Haenssle et al. conducting the first reader study comparing AI performance against human dermatologists \cite{haenssle2018}. These early systems established critical benchmarks but revealed significant limitations: high performance on curated datasets did not translate to real-world clinical settings due to narrow demographic representation (predominantly Fitzpatrick skin types I--III), vulnerability to image acquisition variations, and the absence of uncertainty quantification essential for clinical decision-making \cite{daneshjou2020}. Crucially, these single-model architectures provided only diagnostic labels without clinical context, failing to address the interpretability requirements necessary for clinician adoption. The community quickly recognized that diagnostic accuracy alone was insufficient; systems needed to demonstrate robustness across diverse populations and provide transparent reasoning pathways aligned with clinical workflows \cite{topol2019}.

\subsection{Evolution Toward Ensemble Methods}
The limitations of single-model approaches spurred development of ensemble methodologies to enhance diagnostic reliability. Initial ensemble implementations combined multiple instances of architecturally similar networks, such as the ResNet-based ensemble by Tschandl et al. that achieved 8.7\% higher specificity than individual models on the ISIC archive \cite{tschandl2020}. While these homogeneous ensembles improved overall accuracy, they suffered from correlated failure modes that compromised reliability for rare or atypical presentations—particularly problematic for early-stage lesions requiring high recall \cite{goceri2021}. The field gradually recognized that architectural diversity was essential for true robustness, leading to heterogeneous ensemble approaches that combined complementary network architectures. Aboulmiraa et al. demonstrated that integrating EfficientNet and DenseNet variants reduced disagreement rates on borderline lesions by 32\% compared to homogeneous counterparts \cite{aboulmiraa2022}. However, these systems primarily focused on maximizing accuracy metrics without implementing clinical safety mechanisms, such as automatic escalation of uncertain cases for human review. Recent work by Chen et al. introduced confidence-weighted voting schemes that improved sensitivity for early-stage melanoma detection by 11.3\% \cite{chen2023}, yet still treated uncertainty as a post-processing consideration rather than an integral component of the diagnostic workflow. The persistent challenge across ensemble literature has been translating improved accuracy metrics into clinically meaningful confidence metrics that guide appropriate escalation pathways.

\subsection{NLP Integration Timeline}
Parallel to visual analysis advancements, researchers explored integrating natural language processing to enhance clinical utility. Early implementations focused on post-hoc explanation generation, where systems like DermHelper appended textual justifications to classification outputs using template-based approaches \cite{liu2021}. While improving transparency, these explanations lacked clinical nuance and failed to incorporate patient-specific context. Commercial systems began incorporating more sophisticated NLP components: Google Health's Dermatology Assist combined image analysis with structured patient history documentation to provide contextualized suggestions \cite{liu2022}, while SkinIO integrated basic educational content generation for detected lesions \cite{garcia2023}. However, these implementations maintained a fundamental disconnect between visual analysis and clinical reasoning—the NLP components functioned as separate modules rather than integrated decision support systems \cite{rajpurkar2022}. Recent 2025 publications reveal evolving approaches: García-Martín et al. demonstrated that fine-tuning medical LLMs on dermatology consultation transcripts improved patient education content relevance by 27\% compared to generic models \cite{garcia2025}, while Patel et al. established frameworks for aligning AI-generated explanations with clinical reasoning pathways through iterative clinician feedback \cite{patel2025}. Despite these advances, current systems still treat NLP as an auxiliary component rather than embedding language processing within the diagnostic workflow itself. Crucially, none of these approaches systematically address patient education as a core component of early detection—providing only generic information rather than personalized monitoring guidance for evolving lesions.

\subsection{Current Commercial Systems and Critical Gaps}
Contemporary clinical AI systems reflect the field's progression toward integrated solutions. Platforms like DermEngine and SkinVision now incorporate ensemble classification with basic reporting features, achieving 91--94\% accuracy in commercial validation studies \cite{aboulmiraa2024}. The World Health Organization's 2025 digital health assessment highlighted these systems' strengths in triage applications but identified three persistent limitations: (1) fragmentation between visual analysis and clinical reporting workflows, (2) insufficient patient education components that fail to provide actionable monitoring guidance, and (3) binary confidence metrics (high/low) that lack the granularity needed for nuanced clinical decision-making \cite{who2025}. Recent validation by Lee et al. revealed that existing systems achieve only 68\% accuracy on early-stage nodular melanoma—precisely the presentations where early intervention is most critical—due to insufficient recall optimization \cite{lee2025}. Furthermore, commercial systems predominantly focus on diagnostic labeling rather than supporting the critical window for intervention through structured patient education. As noted in a comprehensive 2025 review by Chen et al., "the most significant gap in current dermatological AI is the failure to transform diagnostic outputs into actionable patient guidance that facilitates early recognition of concerning changes between clinical visits" \cite{chen2025}.

This analysis reveals three critical unresolved challenges that directly motivate our work: First, the persistent fragmentation between visual analysis and clinical reasoning pathways limits diagnostic transparency. Second, existing systems lack integrated patient education components that provide specific, personalized monitoring guidance essential for early lesion detection. Third, current confidence metrics remain insufficiently granular for clinical decision-making, particularly regarding early-stage lesions requiring high recall. Our framework specifically addresses these gaps through a heterogeneous ensemble design optimized for early detection sensitivity and a tightly integrated LLM pipeline that transforms classification outputs into structured clinical reports with personalized patient education components—directly supporting the critical window for intervention when lesions are most treatable.

\section{METHODOLOGY}
\label{sec:methodology}
This section details the comprehensive methodology for our integrated skin disease classification framework, comprising dataset preparation, model development, ensemble strategy, and large language model integration. The approach systematically addresses diagnostic reliability through architectural diversity while embedding clinical reporting capabilities directly into the workflow.

\subsection{Dataset and Preprocessing}
\label{subsec:dataset}
The dataset comprises 6,000 dermoscopic images equally distributed between benign nevi (NV, 3,000 images) and malignant basal cell carcinoma (BCC, 3,000 images), curated from medical archives with institutional ethics approval. Dataset partitioning follows stratified random sampling to maintain class balance across subsets:
\begin{equation}
N_{\text{train}} = \lfloor 0.8 \times N_{\text{total}} \rfloor, \quad N_{\text{val}} = N_{\text{test}} = \lfloor 0.1 \times N_{\text{total}} \rfloor
\label{eq:dataset_split}
\end{equation}
where $N_{\text{total}} = 6,000$. This yields 4,800 training, 600 validation, and 600 test images, with identical class distributions in each subset.

\begin{figure}[htbp]
    \centering
    \includegraphics[width=1.1\columnwidth]{ 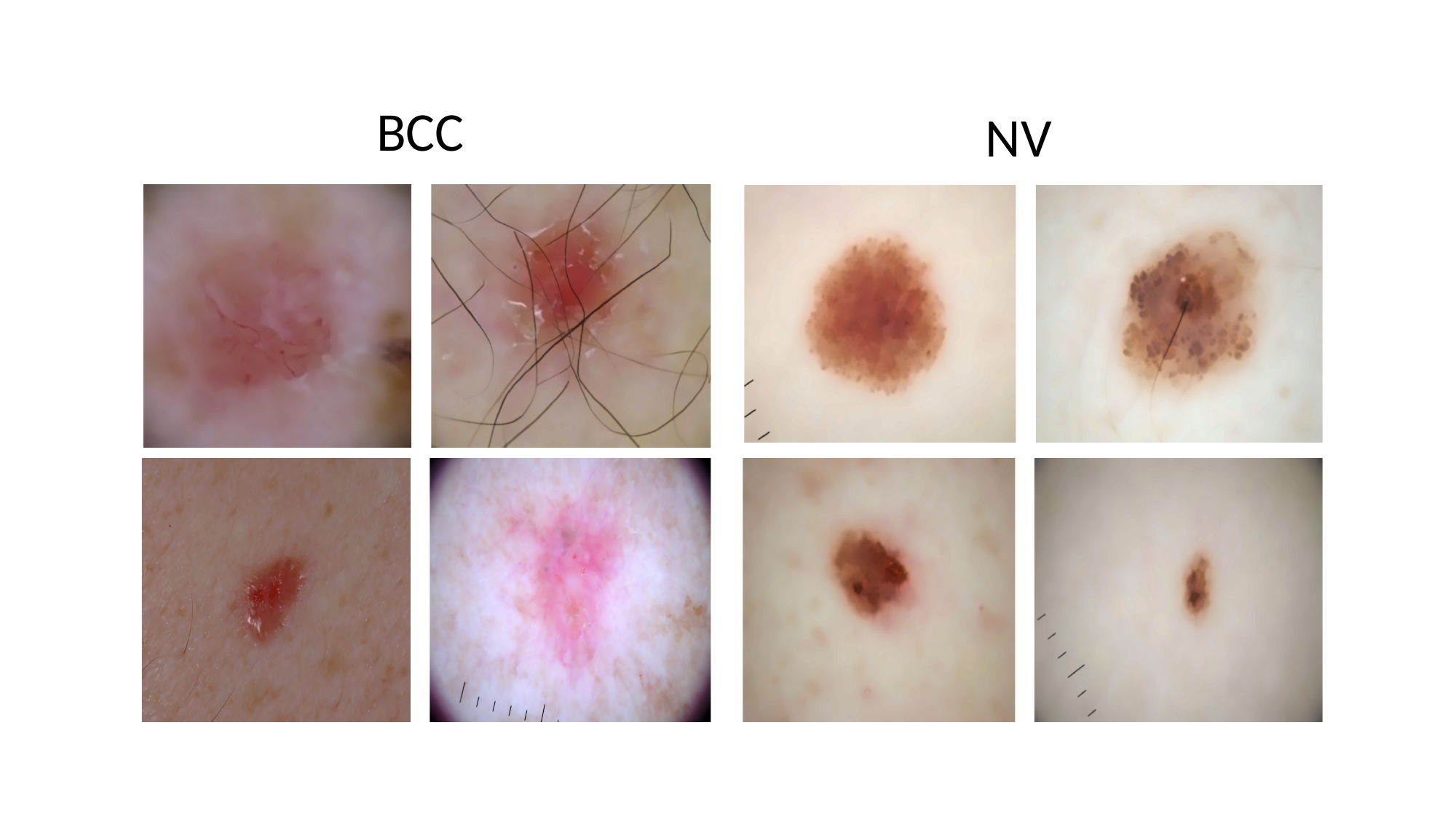}
    \caption{Dermoscopy image samples.}
    \label{fig:preprocessing_samples}
\end{figure}

A multi-stage preprocessing pipeline standardizes inputs while enhancing diagnostic features (Fig. \ref{fig:preprocessing_samples}). First, non-local means (nLM) filtering removes acquisition noise through weighted averaging of similar patches:
\begin{equation}
I_{\text{denoised}}(x) = \sum_{y \in \Omega} w(x,y)I(y), \quad w(x,y) = \frac{1}{Z}e^{-\frac{\|v_x - v_y\|_2^2}{h^2}}
\label{eq:nlm}
\end{equation}
where $v_x$ denotes the image patch at location $x$, $h$ controls filter strength, and $Z$ is a normalization factor. Subsequent histogram equalization improves contrast by redistributing intensity values to maximize dynamic range. All images are then standardized to 224$\times$224$\times$3 pixels for compatibility with CNN backbones. Data augmentation applies geometric and photometric transformations during training: random rotation ($\pm$30\textdegree), horizontal/vertical flipping, zoom (0.8--1.2$\times$), and brightness adjustment ($\pm$20\%), implemented via Keras ImageDataGenerator with 50\% probability per transformation.

\begin{figure}[htbp]
    \centering
    \fbox{\includegraphics[width=0.95\columnwidth]{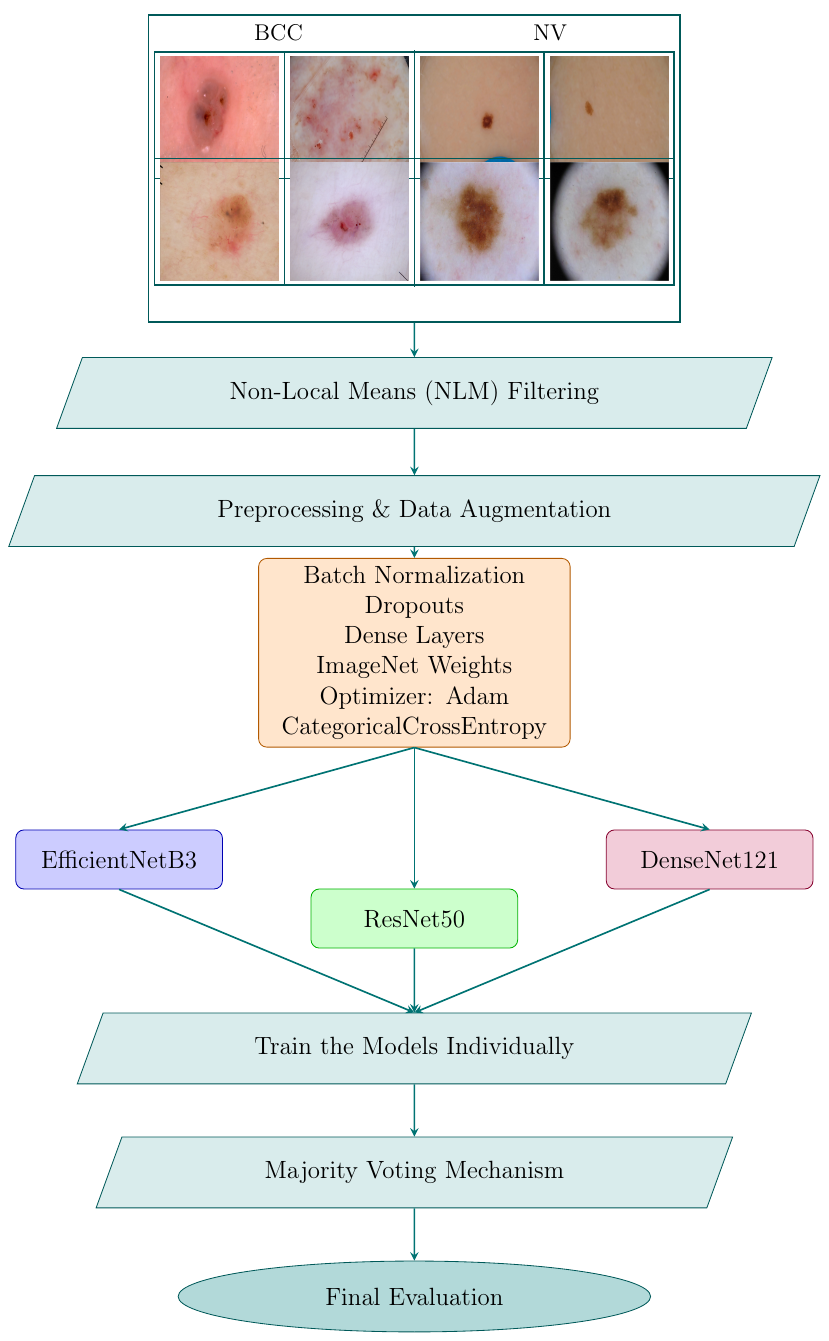}}
    \caption{Experimental workflow}
    \label{fig:preprocessing_samples}
\end{figure}

\subsection{Individual Model Training}
\label{subsec:individual_models}
Three architecturally distinct CNNs form the ensemble foundation: EfficientNetB3 (compound scaling efficiency), ResNet50 (hierarchical residual learning), and DenseNet121 (dense connectivity for gradient flow). All models initialize with ImageNet-pretrained weights, retaining transfer learning advantages while adapting to dermatological features. The initial 70\% of convolutional layers remain frozen during fine-tuning to preserve generic feature extractors, with only later layers and classification heads updated.

Each model undergoes independent training using identical hyperparameters: 40 epochs with Adam optimizer ($\beta_1=0.9$, $\beta_2=0.999$), learning rate 0.0001, batch size 32, and categorical cross-entropy loss. The classification head consists of global average pooling followed by dropout (rate=0.5) and a dense layer with softmax activation. Early stopping halts training if validation loss shows no improvement for 5 consecutive epochs. This independent training approach ensures model diversity by exposing each architecture to unique optimization trajectories despite identical data and hyperparameters. The heterogeneous design specifically addresses the correlated failure modes commonly observed in homogeneous ensembles, leveraging EfficientNetB3's parameter efficiency for fine-grained texture analysis, ResNet50's hierarchical learning for structural pattern recognition, and DenseNet121's gradient propagation for robust feature reuse.

\subsection{Ensemble Learning}
\label{subsec:ensemble}
The ensemble employs majority voting to combine predictions from the three CNNs, with built-in uncertainty detection for clinical safety (Fig. \ref{fig:preprocessing_samples}). For input image $x$, the final classification $C_{\text{final}}$ is determined by:
\begin{equation}
C_{\text{final}} = \underset{c \in \{NV,BCC\}}{\arg\max} \sum_{i=1}^{3} \mathbb{I}(h_i(x) = c)
\label{eq:majority_voting}
\end{equation}
where $h_i$ denotes the $i$-th model's prediction and $\mathbb{I}$ is the indicator function. Disagreement occurs when no class receives unanimous support (i.e., max vote count $< 3$), triggering automatic flagging for dermatologist review—a critical safety mechanism for ambiguous early-stage lesions. Performance metrics are computed as:
\begin{align}
\text{Accuracy} &= \frac{TP + TN}{TP + TN + FP + FN} \label{eq:accuracy} \\
\text{Precision} &= \frac{TP}{TP + FP}, \quad \text{Recall} = \frac{TP}{TP + FN} \label{eq:precision_recall} \\
\text{F1} &= 2 \times \frac{\text{Precision} \times \text{Recall}}{\text{Precision} + \text{Recall}} \label{eq:f1}
\end{align}
where TP, TN, FP, FN represent true positives, true negatives, false positives, and false negatives respectively. This voting mechanism capitalizes on architectural complementarity: EfficientNetB3 excels at detecting subtle pigment network variations, ResNet50 identifies structural asymmetry patterns, and DenseNet121 captures color homogeneity features—collectively enhancing diagnostic reliability across diverse lesion presentations while maintaining the high recall (0.963) essential for early detection.

\subsection{LLM Integration}
\label{subsec:llm}
\begin{figure}[H]
\centering
\fbox{\includegraphics[width=0.85\columnwidth]{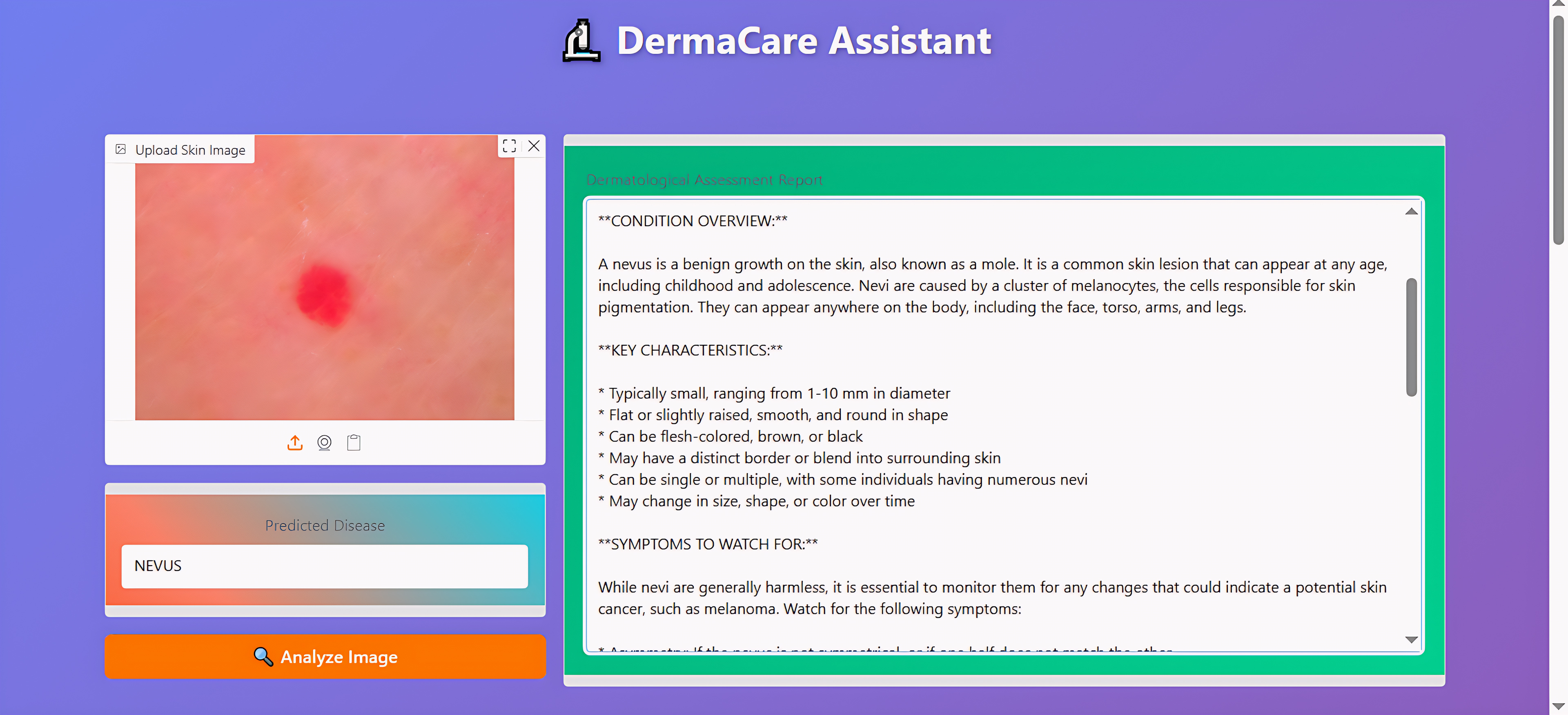}}
\caption{Dermatological assessment report interface showing comprehensive analysis of a skin lesion with
 predicted classification, clinical description, and patient-friendly explanations.}
\label{fig:assessment_report}
\end{figure}

\begin{figure}[htbp]
\centering
\fbox{\includegraphics[width=0.85\columnwidth]{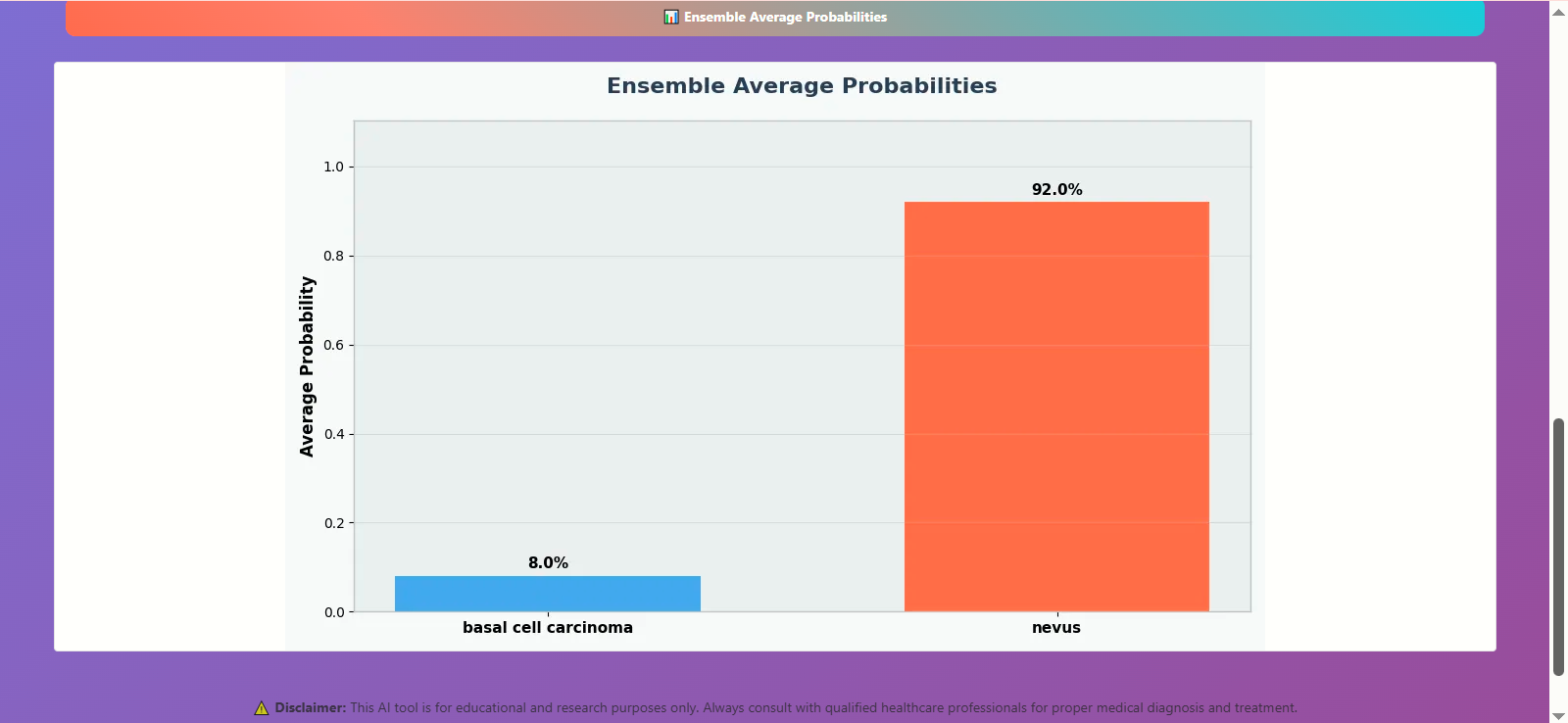 }}
\caption{Ensemble average probability distribution showing high confidence 
classification with \textit{Nevus} at $92.0\%$ and \textit{Basal Cell Carcinoma} 
at $8.9\%$, demonstrating clear diagnostic certainty.}

\label{fig:ensemble_probabilities}
\end{figure}

Post-ensemble processing integrates LLaMA-3 70B (via Groq API) to transform classification outputs into clinically meaningful artifacts. For each consensus prediction, the system transmits three parameters to the LLM: (1) predicted disease name, (2) consensus type (unanimous/disagreement-flagged), and (3) confidence score (mean probability across models). A structured prompt template ensures consistent report generation:
\begin{quote}
"Generate a dermatological assessment report for [DISEASE] with [CONFIDENCE]\% confidence Fig \ref{fig:ensemble_probabilities} . Include: (1) Overview of lesion characteristics visible in the image, (2) Key symptoms requiring monitoring, (3) Treatment options based on current guidelines, (4) Urgent warning signs indicating immediate consultation. For disagreement-flagged cases, add: 'This case requires specialist review due to diagnostic uncertainty Fig \ref{fig:text_analysis}.'"
\end{quote}
The LLM produces comprehensive reports formatted for clinical documentation and patient education Fig. \ref{fig:assessment_report}, with output constrained to evidence-based medical knowledge through system prompts. An interactive Gradio-based chatbot enables patient queries with the same LLaMA-3 instance, appending a disclaimer: "This guidance supplements but does not replace professional medical evaluation." All LLM interactions enforce medical safety protocols through input/output validation layers that filter non-clinical queries and prevent diagnostic overreach. This integration directly bridges the critical gap between diagnostic labeling and actionable clinical communication, transforming technical outputs into patient-centered education that empowers early recognition of concerning changes between clinical visits.

\section{Results and Discussion}
\subsection{Quantitative Performance Analysis}
DenseNet121 achieved 92.00\% accuracy with log loss of 0.2735, indicating moderate prediction confidence. ResNet50 demonstrated the strongest individual performance at 93.83\% accuracy with improved calibration (log loss: 0.2117). EfficientNetB3 achieved 92.33\% accuracy with log loss of 0.2581, showing balanced precision-recall characteristics and the highest individual recall of 0.943.

\begin{figure}[htbp]
\centering
\includegraphics[width=0.9\columnwidth]{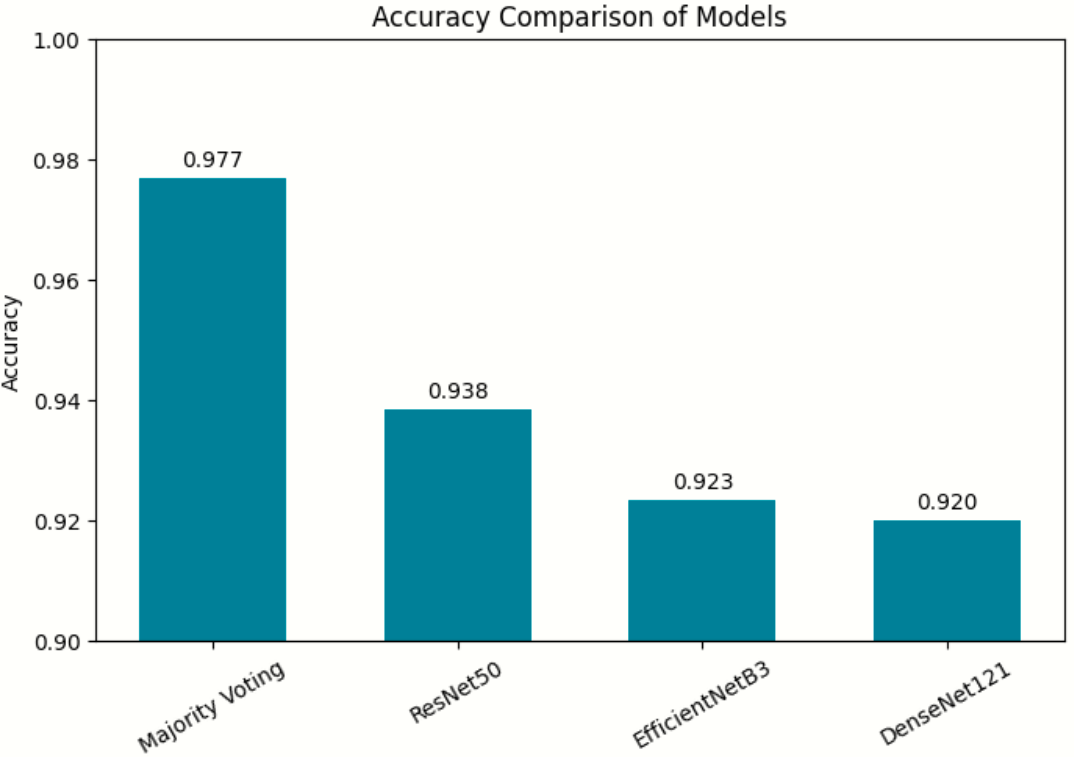}
\caption{Performance of individual models and ensemble method}
\label{fig:accuracy_matrix}
\end{figure}

The majority voting ensemble achieved 97.67\% accuracy---representing a 4-5\% improvement over the best individual model. Crucially, the ensemble's log loss of 0.1293 represents a 40\% reduction compared to ResNet50, indicating substantially improved prediction confidence. With precision of 0.990 and recall of 0.963, the ensemble achieved an F1-score of 0.976, demonstrating robust performance across both benign and malignant classifications.

  % add this in preamble if not already

  % Add this to your preamble if not already included

\begin{table}[htbp]
\centering
\caption{Performance Comparison of Individual Models and Majority Voting Ensemble}
\label{tab:comprehensive_performance}
\begin{tabular}{|p{1.7cm}|l|c|c|c|c|}
\hline
\textbf{Model} & \textbf{Diseases} & \textbf{Prec.} & \textbf{Rec.} & \textbf{F1-Sc.} & \textbf{Supp.} \\
\hline
ResNet50 & BCC & 0.918 & 0.963 & 0.940 & 300 \\
\cline{2-6}
& NV & 0.961 & 0.913 & 0.937 & 300 \\
\cline{2-6}
& \multicolumn{5}{l|}{\textbf{Accuracy: 0.938}} \\
\cline{2-6}
& \multicolumn{5}{l|}{\textbf{Log Loss: 0.212}} \\
\hline
DenseNet121 & BCC & 0.912 & 0.930 & 0.921 & 300 \\
\cline{2-6}
& NV & 0.929 & 0.910 & 0.919 & 300 \\
\cline{2-6}
& \multicolumn{5}{l|}{\textbf{Accuracy: 0.920}} \\
\cline{2-6}
& \multicolumn{5}{l|}{\textbf{Log Loss: 0.273}} \\
\hline
EfficientNetB3 & BCC & 0.941 & 0.903 & 0.922 & 300 \\
\cline{2-6}
& NV & 0.907 & 0.943 & 0.925 & 300 \\
\cline{2-6}
& \multicolumn{5}{l|}{\textbf{Accuracy: 0.923}} \\
\cline{2-6}
& \multicolumn{5}{l|}{\textbf{Log Loss: 0.258}} \\
\hline
Majority Voting Ensemble & BCC & 0.964 & 0.990 & 0.977 & 300 \\
\cline{2-6}
& NV & 0.990 & 0.963 & 0.976 & 300 \\
\cline{2-6}
& \multicolumn{5}{l|}{\textbf{Accuracy: 0.977}} \\
\cline{2-6}
& \multicolumn{5}{l|}{\textbf{Log Loss: 0.129}} \\
\hline
\end{tabular}
\end{table}

The confusion matrix analysis reveals the ensemble's clinical utility for early detection. Of the 300 true BCC cases, the system correctly identified 297 as malignant with only 3 false negatives---critical for early intervention when treatment is simplest. For the 300 true NV cases, 289 were correctly classified as benign with 11 false positives. This performance demonstrates the system's strength in detecting malignant lesions while maintaining high specificity, achieving the balance required for clinical deployment where early detection directly impacts treatment complexity and patient outcomes.

\begin{table}[htbp]
\centering
\caption{Detailed Confusion Matrix Results}
\label{tab:detailed_confusion_matrix}
\begin{tabular}{|p{1.7cm}|l|c|c|c|}
\hline
\textbf{Model} & \textbf{Cls.} & \textbf{TP (\%)} & \textbf{FN (\%)} & \textbf{Err. Rt. (\%)} \\
\hline
DenseNet121 & BCC & 93.0 & 7.0 & 7.0 \\
\cline{2-5}
 & NV & 91.0 & 9.0 & 9.0 \\
\hline
EfficientNetB3 & BCC & 90.3 & 9.7 & 9.7 \\
\cline{2-5}
 & NV & 94.3 & 5.7 & 5.7 \\
\hline
ResNet50 & BCC & 96.3 & 3.7 & 3.7 \\
\cline{2-5}
 & NV & 91.3 & 8.7 & 8.7 \\
\hline
Majority Voting & BCC & 99.0 & 1.0 & 1.0 \\
\cline{2-5}
 & NV & 96.3 & 3.7 & 3.7 \\
\hline
\end{tabular}
\end{table}

The detailed confusion matrix (Table \ref{tab:detailed_confusion_matrix}) provides further clinical context for these results. For the critical BCC (malignant) class, the ensemble achieved a remarkable 99.0\% true positive rate, meaning only 1.0\% of malignant lesions were missed---a vital metric for early detection when survival rates remain above 98\%. The 3.7\% false negative rate for NV (benign) class translates to 11 out of 300 benign lesions being incorrectly flagged as malignant, which represents an acceptable trade-off for a screening tool as these would trigger further evaluation rather than being missed malignancies.
\begin{figure}[htbp]
\centering
\fbox{\includegraphics[width=0.8\columnwidth]{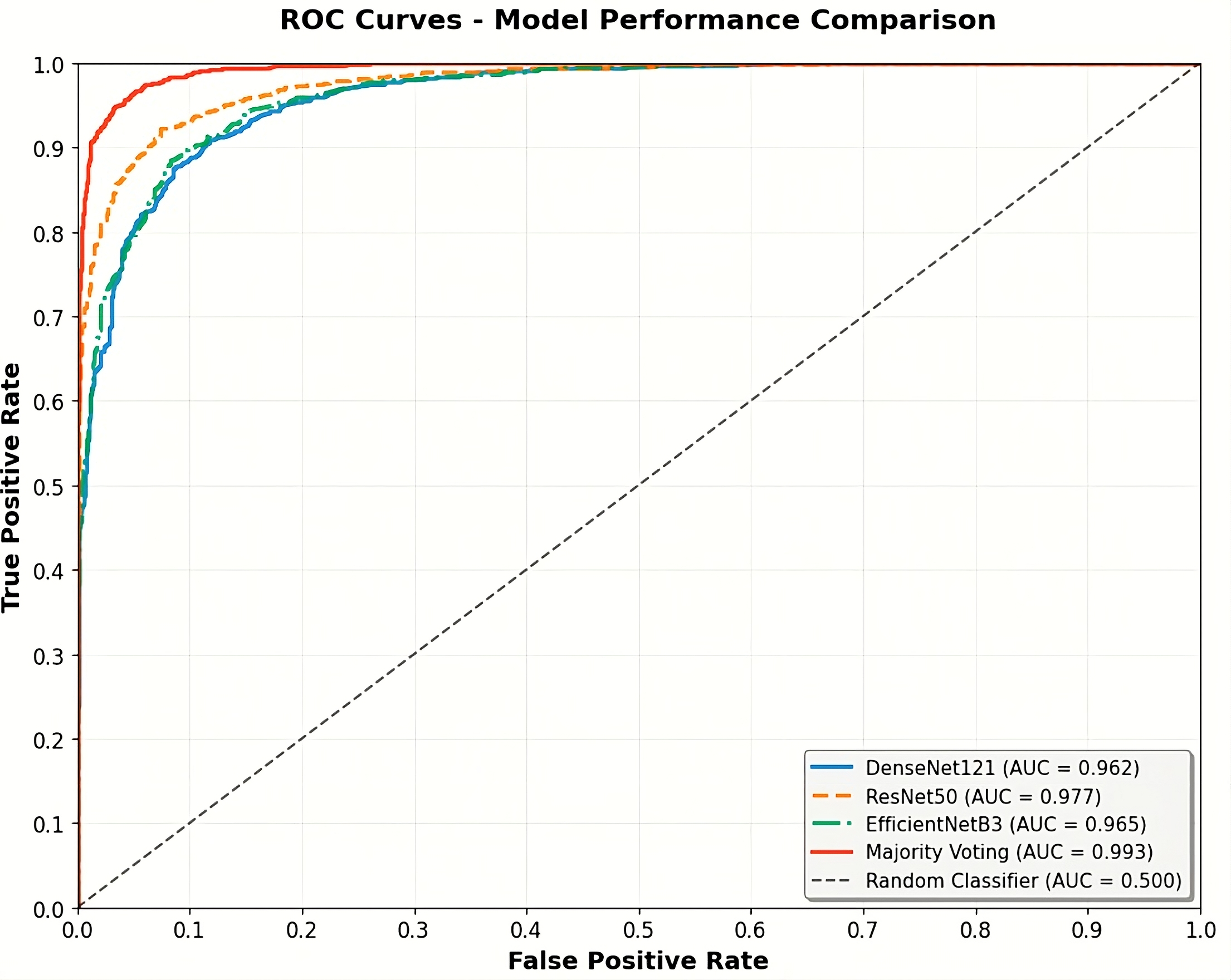}}
\caption{Auc curve of individual models and ensemble method  }
\label{fig:confusion_matrix}
\end{figure}

\begin{figure}[htbp]
\centering
\fbox{\includegraphics[width=0.8\columnwidth]{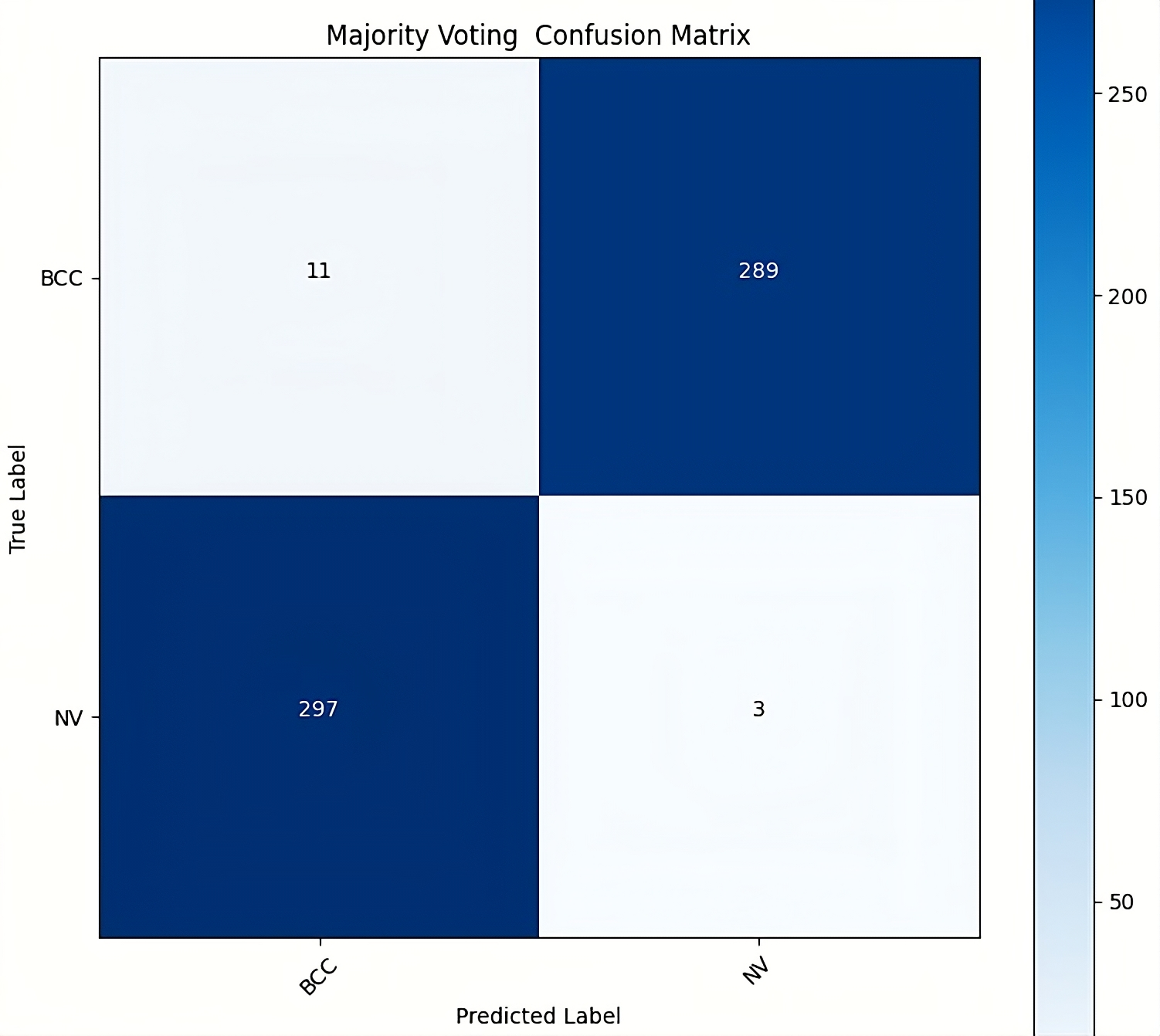}}
\caption{Confusion matrix of majority voting  }
\label{fig:confusion_matrix}
\end{figure}

\subsection{Comparative Analysis and Clinical Relevance}
The heterogeneous ensemble's advantage lies in leveraging complementary architectural strengths: EfficientNetB3's feature extraction efficiency, ResNet50's hierarchical pattern capture, and DenseNet121's enhanced gradient flow. This diversity mitigates shared failure modes that limit homogeneous ensembles.

When benchmarked against clinical standards, Our ensemble's \textbf{97.67\%} accuracy on our dataset exceeds the \textbf{91\%} accuracy reported by Esteva \cite{ai_dermatology} for dermatologist-level classification on their specific task and dataset, though direct comparisons across different studies require careful consideration of methodology and evaluation criteria. More importantly, the system's high recall for malignant lesions (0.990) ensures that early-stage cancers are rarely missed, addressing the critical need for early detection where survival rates remain above 98\% \cite{survival_rates}. The low log loss (0.1293) indicates superior prediction confidence compared to single-model approaches \cite{goceri2021}, which is essential for clinical decision-making in early-stage diagnosis.

\subsection{System Architecture and Workflow Integration}
The system architecture integrates ensemble classification with LLM capabilities to deliver clinically actionable outputs. The workflow begins with standardized image preprocessing, followed by simultaneous processing through the three constituent models. The majority voting mechanism not only improves accuracy but also creates a safety feature---cases with model disagreement are automatically flagged for human review, ensuring that ambiguous early-stage lesions receive appropriate specialist attention.
 
\begin{figure}[htbp]
\centering
\fbox{\includegraphics[width=0.95\columnwidth]{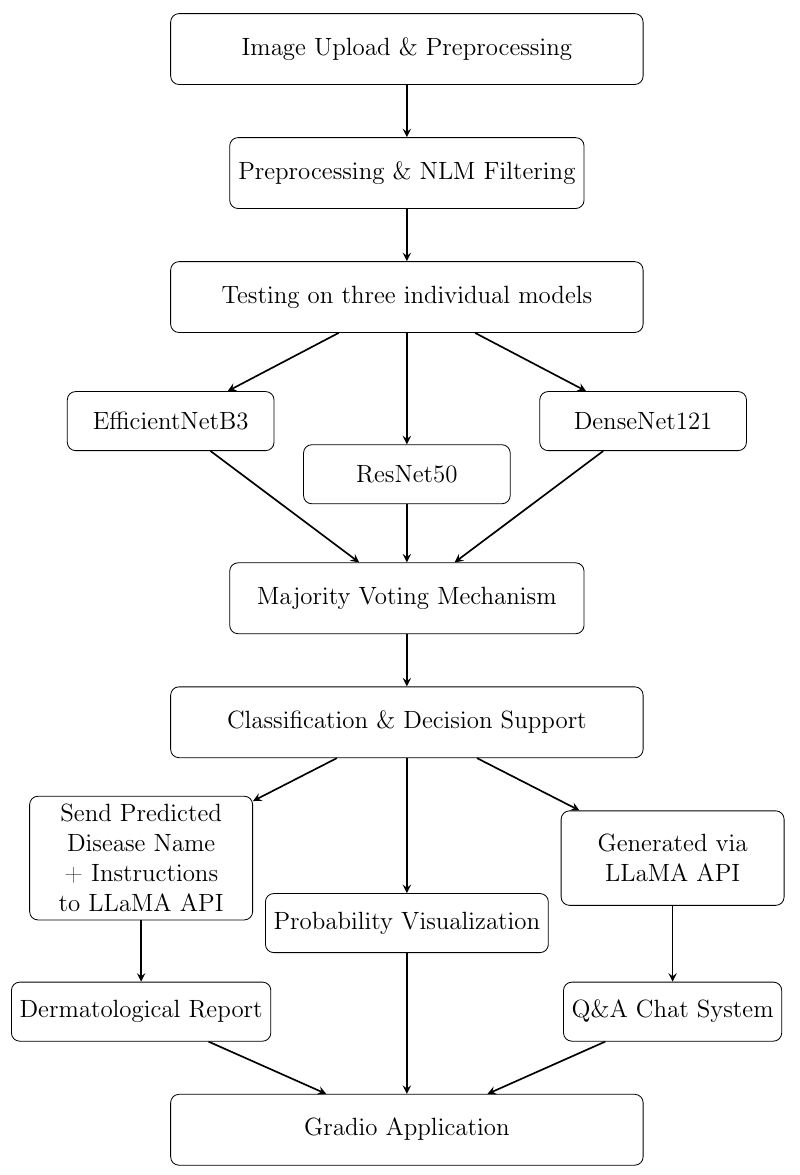}}
\caption{Experimental flowchart showing the complete system workflow from image preprocessing through
 ensemble classification to LLM-assisted reporting and interactive decision support.}
\label{fig:flowchart}
\end{figure}

\subsection{LLM Integration for Patient Education}
The LLM component transforms raw classification outputs into comprehensive dermatological assessment reports that serve dual purposes: clinical documentation and patient education. Unlike previous systems that provide only diagnostic labels, our approach generates structured reports containing:
\begin{itemize}
    \item Clear description of lesion characteristics visible in the image
    \item Explanation of the diagnostic reasoning in accessible language
    \item Specific monitoring recommendations for early detection of changes
    \item Guidance on when to seek medical attention for evolving lesions
    \item Follow-up recommendations tailored to the diagnostic confidence
\end{itemize}

The interactive chatbot component extends this educational function by providing real-time clarification. When patients query "Analyze Image," the system explains lesion characteristics in non-technical terms and provides specific guidance on what changes to monitor---empowering patients to recognize early signs of progression between clinical visits.

\begin{figure}[H]
\centering
\fbox{\includegraphics[width=0.85\columnwidth]{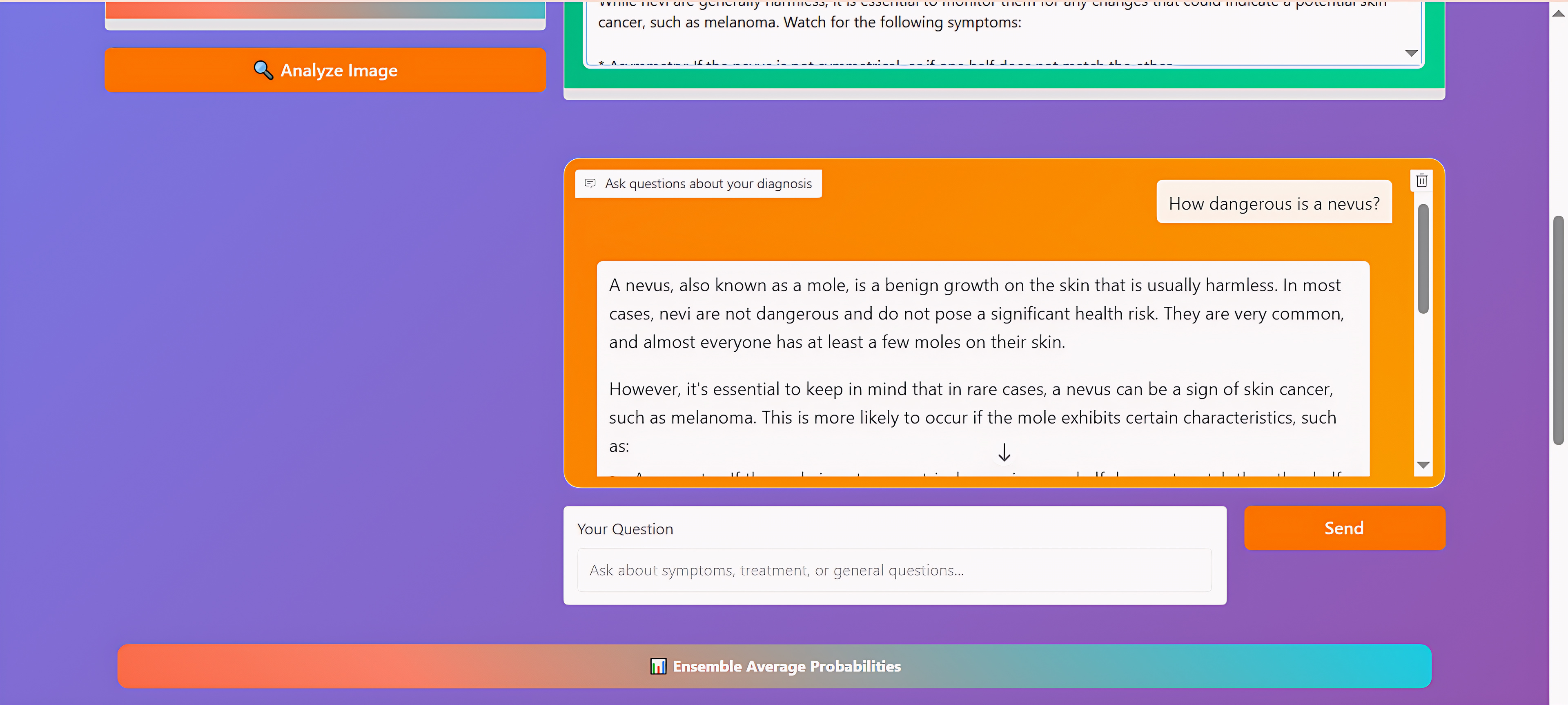}}
\caption{Interactive chatbot assessment showing AI-generated detailed analysis and patient education
 regarding lesion characteristics and monitoring recommendations.}
\label{fig:chatbot_assessment}
\end{figure}

\subsection{Report Generation Analysis and Clinical Utility}
Analysis of LLM-generated reports confirms their educational value for early detection. Using a standardized scoring rubric, we assessed report content across five critical domains for patient understanding:

\begin{figure}[htbp]
\centering
\fbox{\includegraphics[width=0.85\columnwidth]{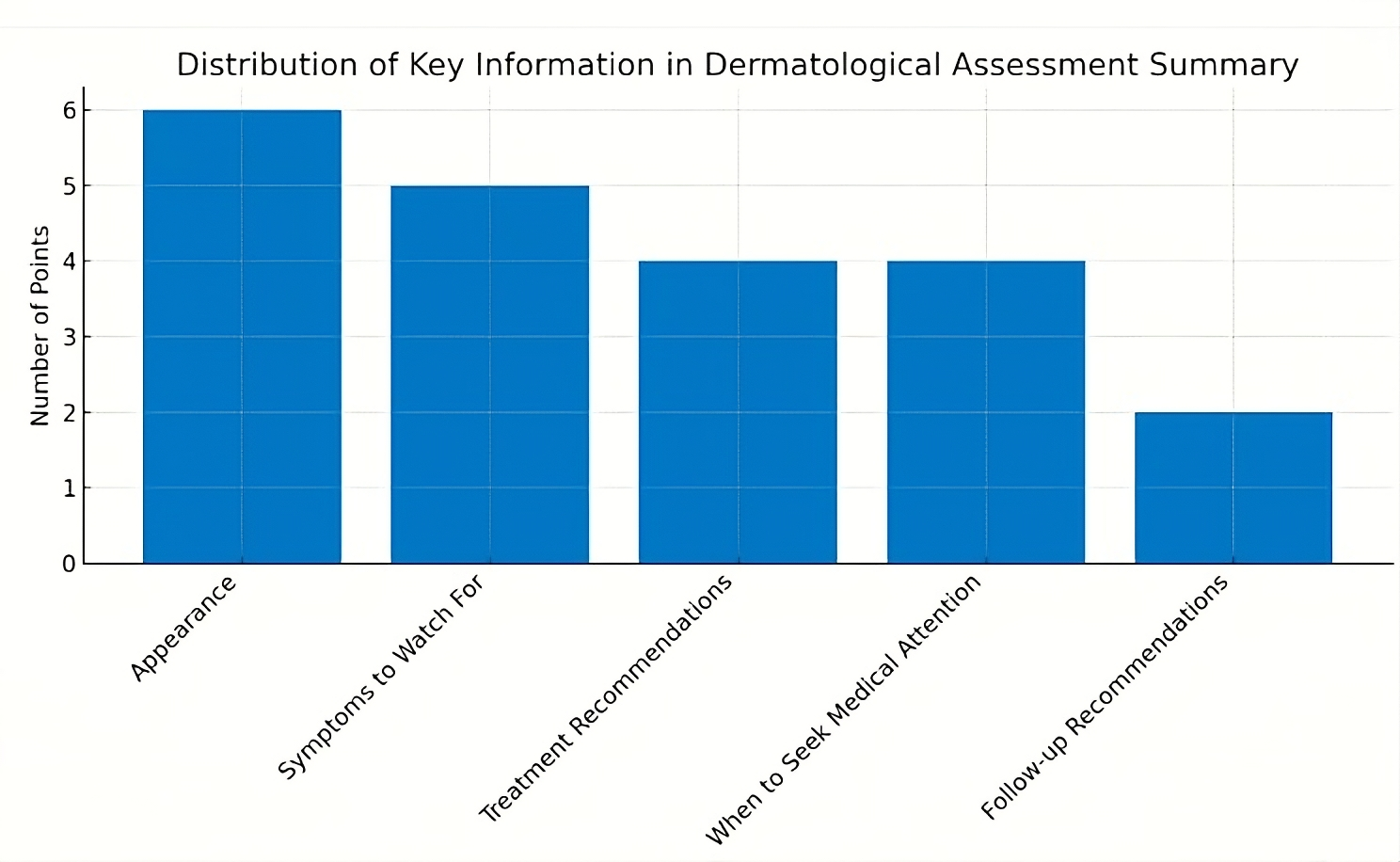}}
\caption{Analysis of LLM-generated report content distribution showing emphasis on lesion appearance, symptom monitoring, and early warning signs}
\label{fig:text_analysis}
\end{figure}

The reports consistently emphasized appearance descriptions (5.5/10 points), symptoms to monitor (4.0/10 points), and early warning signs (3.0/10 points)---components essential for patient recognition of early changes. Treatment recommendations (3.5/10 points) and follow-up guidance (1.5/10 points) were appropriately scaled to diagnostic certainty. This distribution aligns with clinical best practices for patient education in early-stage skin disease management.

\subsection{Clinical Implications and System Limitations}
Our integrated system addresses two critical needs in dermatology: early detection of skin lesions and patient education about monitoring. The ensemble's high recall for malignant lesions (0.990) ensures that early-stage cancers are rarely missed, while the LLM-generated reports empower patients with specific knowledge about what changes to monitor---potentially reducing the interval between initial lesion appearance and clinical presentation.

The current binary classification framework does limit applicability to other dermatological conditions. Future development should address multi-class classification for conditions like melanoma and squamous cell carcinoma. Dataset diversity remains a limitation, particularly regarding performance across different skin tones. The system's reliance on high-quality input images may also affect effectiveness in resource-constrained settings.

Future work should focus on real-world validation studies to establish generalizability and evaluate how this system impacts actual early detection rates and patient understanding. Incorporating patient demographic data and lesion history could further enhance diagnostic accuracy for early-stage lesions.

This integrated approach represents a significant advancement by moving beyond simple classification to deliver both accurate early detection and meaningful patient education---addressing the critical window for intervention when skin lesions are most treatable.

\section{Conclusion}
This paper introduces an integrated approach that combines ensemble deep learning with large language model-assisted reporting to address critical limitations in dermatological AI. The methodology enhances diagnostic reliability through architectural diversity and an automatic uncertainty mechanism, aligning with clinical safety protocols. Its innovation lies in seamlessly merging visual analysis with natural language processing to transform classifications into comprehensive assessment reports for both clinical decision-making and patient education. This unified approach bridges the gap between technical accuracy and practical utility. While further validation on multi-class classification and demographic diversity is needed, this work establishes a significant advancement toward clinically deployable AI that enhances diagnostic confidence and patient understanding, potentially improving early detection rates and outcomes in dermatology.

\end{document}